\documentclass{article}

\usepackage[preprint,nonatbib]{neurips_2025}

\usepackage[utf8]{inputenc} 
\usepackage[T1]{fontenc}    
\usepackage{hyperref}       
\usepackage{url}            
\usepackage{booktabs}       
\usepackage{amsfonts}       
\usepackage{nicefrac}       
\usepackage{microtype}      
\usepackage{xcolor}         
\usepackage[pdftex]{graphicx}
\usepackage{multirow}
\usepackage{fancyvrb}
\usepackage[ruled,vlined]{algorithm2e}
\usepackage{amsmath}

\makeatletter
\renewcommand{\@noticestring}{} 
\makeatother

\title{TotalFM: An Organ-Separated 3D-CT Foundation Model Leveraging Large-Scale Routine Clinical Radiology Data}

\author{%
  Kohei Yamamoto \\
  Department of Radiology, Jichi Medical University\\
  \texttt{yamamoto.kohei@jichi.ac.jp} \\
  \And
  Tomohiro Kikuchi \\
  Data Science Center, Jichi Medical University\\
  \texttt{r1419kt@jichi.ac.jp} \\
}

\begin{document}

\maketitle

\begin{abstract}
While foundation models in radiology are expected to be applied to various clinical tasks, computational cost constraints remain a major challenge when training on 3D–CT volumetric data. In this study, we propose \textbf{TotalFM}, a radiological foundation model that efficiently learns the correspondence between 3D–CT images and linguistic expressions based on the concept of organ separation, utilizing a large–scale dataset of 140,000 series. By automating the creation of organ volume and finding–sentence pairs through segmentation techniques and Large Language Model (LLM)–based radiology report processing, and by combining self–supervised pre–training via VideoMAE with contrastive learning using volume–text pairs, we aimed to balance computational efficiency and representation capability. In zero–shot organ–wise lesion classification tasks, the proposed model achieved higher F1 scores in 83\% (5/6) of organs compared to CT–CLIP and 64\% (9/14) of organs compared to Merlin. These results suggest that the proposed model exhibits high generalization performance in a clinical evaluation setting using actual radiology report sentences. Furthermore, in zero–shot finding–wise lesion classification tasks, our model achieved a higher AUROC in 83\% (25/30) of finding categories compared to Merlin. We also confirmed performance comparable to existing Vision–Language Models (VLMs) in radiology report generation tasks. Our results demonstrate that the organ–separated learning framework can serve as a realistic and effective design guideline for the practical implementation of 3D–CT foundation models. The source code and pretrained models are publicly available at \url{https://github.com/jichi-labo/TotalFM}
\end{abstract}

\section{Introduction}
\begin{figure}[ht]
  \centering
  \includegraphics[width=1.0\textwidth]{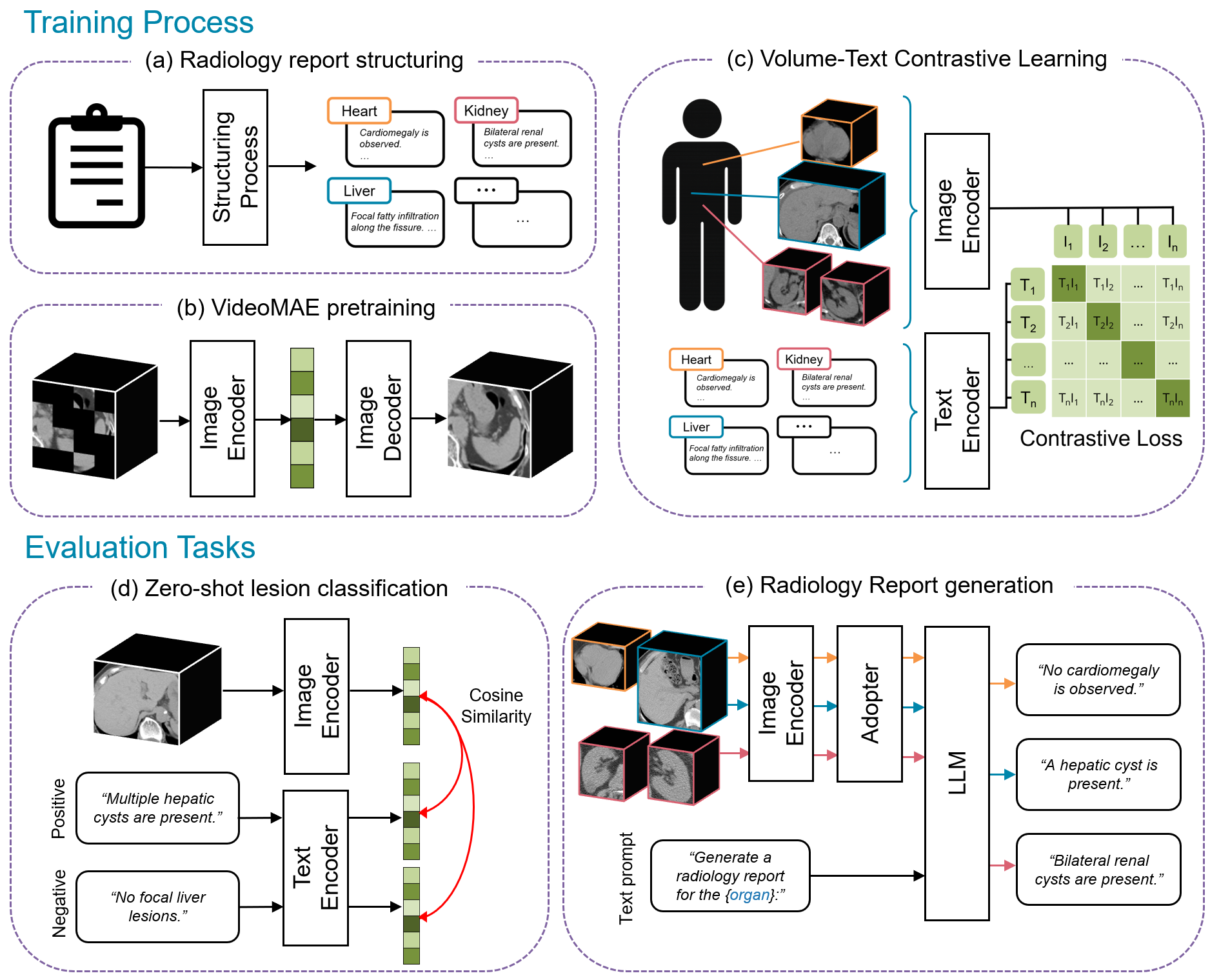}
  \caption{
  \textbf{Overview of the proposed organ–separated framework for 3D–CT vision foundation models.}
  The proposed framework consists of the following steps:
  (a) structuring CT volumes and radiology reports into organ–level units using TotalSegmentator and large language models (LLMs);
  (b) self–supervised pre–training of the image encoder using VideoMAE on a large–scale CT dataset (J-MID);
  (c) organ–wise contrastive learning between segmented CT volumes and corresponding structured report sentences;
  (d) evaluation via zero–shot classification tasks at both organ and finding levels; and
  (e) construction of a vision––language model by connecting the image encoder and an LLM using a Q–Former and LoRA.
  }
  \label{fig:research_overview}
\end{figure}

Foundation models have achieved remarkable success across a wide range of domains by learning general–purpose representations from large–scale data through self–supervised learning, enabling flexible transfer to diverse downstream tasks\cite{Touvron2023-si, Li2023-vr}. In particular, contrastive learning–based approaches that align images and language in a shared embedding space, as exemplified by Contrastive Language–Image Pre–training (CLIP), have enabled flexible image understanding via natural language and have become a central component of vision–language models\cite{Radford2021-ro}. This paradigm has also extended to the medical imaging domain, where foundation models trained on large amounts of unlabeled data have demonstrated strong performance even in zero–shot and few–shot settings. Within medical imaging, volumetric modalities such as CT and MRI are of particular importance, as they contain substantially richer information than 2D images and are critical for comprehensive diagnosis. Effectively leveraging such data therefore requires image encoders that can directly model three–dimensional structures and spatial context.

In contrastive learning, the InfoNCE loss is known to enhance training effectiveness by utilizing non–paired elements within a batch as negative examples, making the acquisition of large batch sizes a key factor in improving accuracy \cite{van-den-Oord2018-pm, Chen2020-du}. However, in the training of 3D foundation models, 3D encoders are computationally expensive, making it extremely difficult to achieve the large batch sizes that dictate contrastive learning performance. For instance, CT–CLIP adopts high spatial resolution, but the associated computational cost limits the feasible batch size to 2 batches per GPU (batch/GPU) \cite{Hamamci2024-xm}. RadFM jointly trains the image encoder and a large language model (LLM), which constrains the batch size to 1 batch/GPU after resizing the input images \cite{Wu2023-jg}. In contrast, Merlin adopts a design that favors larger batch sizes by operating at a coarser spatial resolution \cite{Blankemeier2024-hk}. These design choices reflect a common trade–off in current 3D foundation models between spatial resolution and batch efficiency, which motivates the exploration of alternative strategies for balancing computational cost and representational fidelity.

In this study, we propose an organ–separated framework that balances computational efficiency with high spatial resolution by encoding CT volumes on an organ–by–organ basis as illustrated in Fig.~\ref{fig:research_overview}. This design is inspired by the clinical reading process in which radiologists first assess individual organs before integrating findings into a holistic diagnosis. By decomposing whole–body CT volumes into anatomically meaningful units, the proposed approach aims to mitigate the inherent trade–off between spatial resolution and batch efficiency in existing 3D foundation models. In summary, our contributions to generarist foundation models are as follows:

\begin{itemize}
    \item \textbf{Organ–level volume––text representation learning:}
    We introduce an organ–separated contrastive learning framework that enables efficient learning from high–resolution 3D–CT volumes.

    \item \textbf{Scalable and efficient training of 3D foundation models:}
    By adopting an organ–separated, patch–based design, our approach substantially reduces computational cost and GPU memory usage, enabling large–batch contrastive learning while achieving performance comparable to or exceeding existing 3D foundation models.

    \item \textbf{Automated large–scale data construction:}
    We develop a fully automated pipeline that integrates TotalSegmentator and LLMs to generate approximately 340{,}000 organ–level volume––text pairs from 140{,}000 CT series.
\end{itemize}

\section{Related Works}

\subsection{Current State of 3D Medical Multimodal Foundation Models}
CT–CLIP (2024) is a pioneering 3D contrastive learning model that utilized the ``CT–RATE'' dataset, consisting of approximately 26,000 chest CT scans (around 50,000 volumes after reconstruction) \cite{Hamamci2024-xm}. While it processes 3D volumes directly, large–scale 3D contrastive learning is severely limited by memory constraints. In its public implementation, a batch size of 8 is recommended even in an A100 (80GB) environment, making batch size scaling a practical bottleneck.

RadFM (2023) was proposed as a multimodal foundation model integrating various medical images, including 3D–CT, with LLMs \cite{Wu2023-jg}. While its image encoder is relatively efficient at approximately 81 GFLOPs, it is important to note that input CT images are typically resized from a resolution of $512 \times 512$ down to $256 \times 256$. Furthermore, RadFM adopts a configuration where the image encoder and LLM are trained simultaneously, which restricts the mini–batch size to 1 per GPU due to computational constraints. Consequently, scaling the batch size and utilizing high–resolution information–––both crucial for effective contrastive learning–––remain significant challenges.

M3D (2024) presented a multimodal LLM (MLLM) framework combining 3D Vision Transformers, Perceivers, and LLMs to comprehensively handle tasks such as report generation and Visual Question Answering (VQA) \cite{Bai2024-nb}. However, many existing methods that process the entire 3D volume with a single encoder still face a trade–off between resolution and batch size, leaving room for improvement in extracting localized, subtle findings. 

Pillar–0 (2025) points out that existing industry models often prioritize computational efficiency to the extent of downsampling CT bit depth (12–16 bit) to 8–bit, thereby losing subtle contrast information \cite{Agrawal2025-fy}. Pillar–0 aims to improve generalization performance for radiological images by conducting large–scale pre–training across multiple regions and modalities, including abdomen/pelvis, chest, and head CT, as well as breast MRI.

\subsection{Self–Supervised Learning (SSL): Application of 3D–MAE}
Masked Image Modeling (MIM) and Masked Autoencoders (MAE) have gained attention as powerful self–supervised pre–training methods for 3D medical images \cite{He2022-ti, Zhou2022-qe}. VideoMAE (2023) is a Masked Autoencoder that leverages the spatio–temporal redundancy inherent in video data \cite{Tong2022-hq}. The concept of treating the z–axis in CT as a continuous structure or a pseudo–temporal axis is highly compatible with the pre–training of 3D–CT models. Indeed, representation learning applying MAE to 3D medical images has been reported to contribute to faster training convergence and improved accuracy in downstream tasks. In this study, we implement self–supervised pre–training based on the VideoMAE scheme to acquire fundamental feature extraction capabilities from large–scale unlabeled data.

\subsection{Theory of Maintaining Local Resolution and Patch–based Analysis}
Compared to methods that process whole–body scan data at once, the strategy of partitioning and analyzing regions with anatomical meaning (Region/Organ) is emerging as a key to balancing computational efficiency and resolution. GigaPath (2024), which handles gigapixel–scale pathology images, integrates overall context with localized high–resolution information by dividing images into numerous tiles or patches and processing them hierarchically \cite{Xu2024-as}. This patch/partition–based concept is becoming an essential strategy in medical diagnosis where resolution cannot be sacrificed. Recently, with the advent of high–precision automatic segmentation technologies like TotalSegmentator, it has become feasible to extract organ regions and concentrate computational resources on regions of interest \cite{Wasserthal2023-mc}.

\subsection{Overcoming the Limitations of Whole–Body 3D Encoding}
As shown in Table \ref{tab:related_works_comparison}, most existing 3D medical imaging models face constraints in resolution and computational volume due to their design of encoding the entire body as a single unit. The Organ–Separated Framework proposed in this study extracts organ–specific patches ($192 \times 192 \times 32$) and limits the training target to regions of interest, thereby significantly reducing computations on unnecessary background areas. In preliminary experiments using our implementation environment (H100 GPU, bf16 precision, with controlled organ patch counts), we confirmed that a high mini–batch efficiency of 32 batch/GPU is achievable. This method possesses a clear advantage over existing approaches by enabling large–scale contrastive learning with realistic computational resources while maintaining spatial resolution and information density.

\begin{table}[htbp]
  \centering
  \caption{Comparison of input resolution, GPU type, and batch size among related works.}
  \label{tab:related_works_comparison}
  \begin{tabular}{lcccccc}
    \toprule
    Model & Input Resolution & GFLOPs & GPU Type & GPUs & Batch Size & Batches Per GPU \\
    \midrule
    CT–CLIP & $480 \times 480 \times 240$ & 484 & A100  & 4  & 8  & 2  \\
    RadFM   & $256 \times 256 \times 64$  & 81  & A100  & 32 & 32 & 1  \\
    Merlin  & $224 \times 224 \times 160$ & 667 & A6000 & 1  & 18 & 18 \\
    Pillar–0& $384 \times 384 \times 384$ & 14{,}926 & H100 & 8  & 64 & 8  \\
    \midrule
    TotalFM & $192 \times 192 \times 32$  & 180 & H100  & 2  & 64 & 32 \\
    \bottomrule
  \end{tabular}
\end{table}

\section{Methods}

\subsection{Dataset}
This study was approved by the Institutional Review Board of Jichi Medical University Hospital. In this work, we utilized CT examination data and corresponding Japanese radiology reports from the year 2024, extracted from the Japan Medical Image Database (J-MID) \cite{Akashi2025-xu}. Approximately 280,000 CT series were used for self–supervised pre–training via VideoMAE, and approximately 140,000 series with available reports were used for contrastive learning. 
The training and validation sets were split based on examination dates; patients included in the VideoMAE pre–training or the training sets were strictly excluded from the validation set to prevent data leakage. For the test set, we utilized external data from a single institution that was not included in either the VideoMAE pre–training or the contrastive learning datasets.

For all CT data, organ segmentations were generated using TotalSegmentator and saved in NIfTI format \cite{Wasserthal2023-mc}. Due to the massive scale of the dataset, inference was performed in "fast" mode (half the standard resolution) to ensure computational efficiency.

\subsection{Volume–Text Pair Construction Pipeline}
\begin{figure}[ht]
  \centering
  \includegraphics[width=1.0\textwidth]{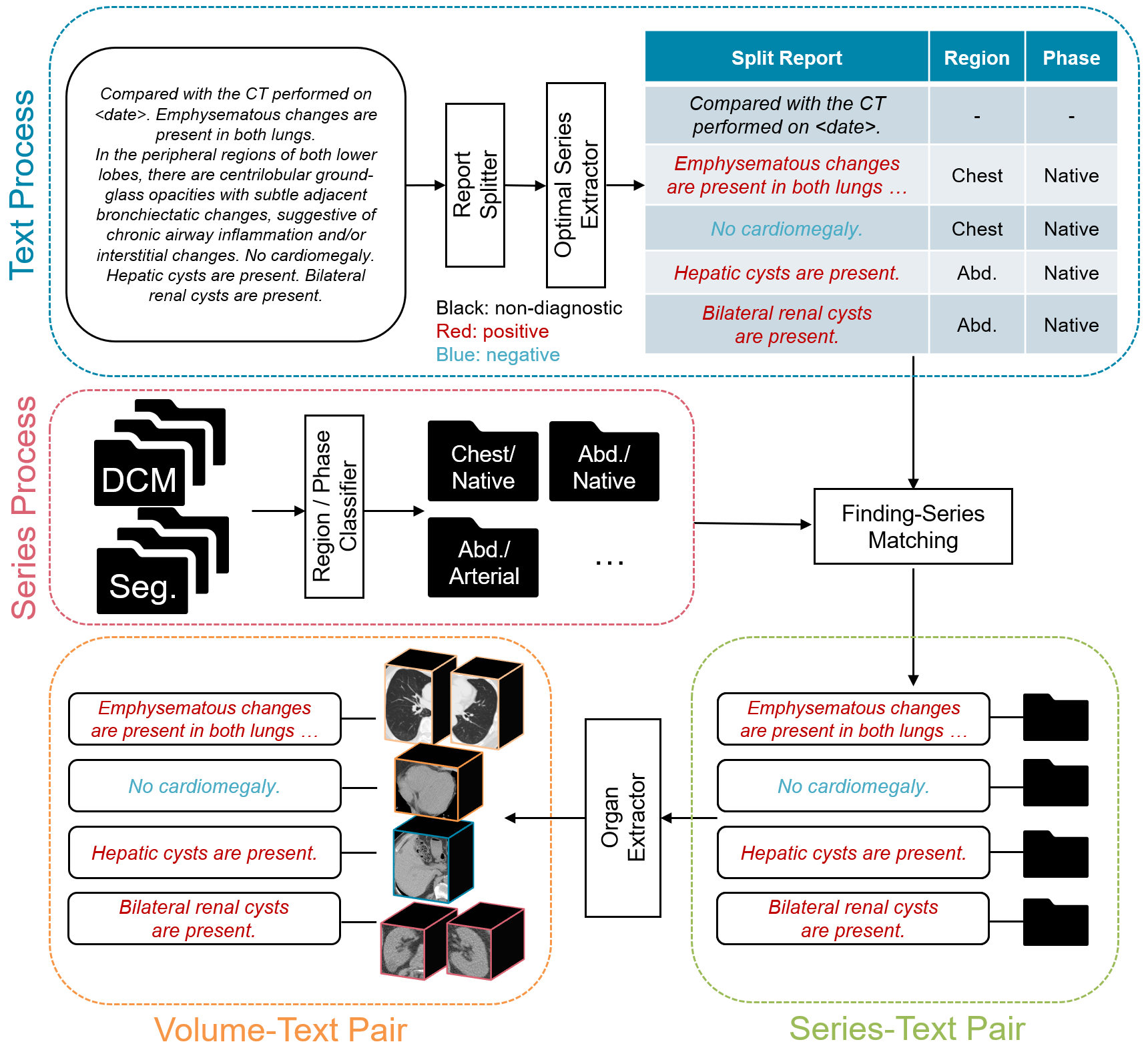}
  \caption{
    Overview of the Volume–Text Pair Construction Pipeline. DCM indicates DICOM data containing images and metadata. Seg. denotes organ segmentations generated by TotalSegmentator.
  }
  \label{fig:dataset_structuring}
\end{figure}

To facilitate organ–separated contrastive learning, we developed a data processing pipeline that decomposes CT volumes and radiology reports into organ–level volume––text pairs, as illustrated in Figure~\ref{fig:dataset_structuring}. The inputs to this pipeline include radiology reports, image data, Digital Imaging and Communications in Medicine (DICOM) metadata, as well as the corresponding TotalSegmentator organ segmentations and contrast phase labels. The pipeline consists of five sequential steps, described below.

\textbf{Step 1: Report Splitter.}
The Report Splitter takes the original radiology report text as input, filters out non–diagnostic sentences, splits the text into individual findings, and determines the presence or absence of disease (positive/negative) for each finding. This study adapts the method developed by Kikuchi et al.~\cite{UnknownUnknown-at}.

\textbf{Step 2: Optimal Series Extractor.}
For each finding sentence, the Optimal Series Extractor predicts the most appropriate imaging region and contrast phase for observation. This step is implemented using prompt–based classification with a LLM (gpt-oss:20b)\cite{openai2025gptoss120bgptoss20bmodel}. The specific prompt is provided in Appendix~\ref{app:llm_prompts:optimal_series_extractor}.

\textbf{Step 3: Region/Phase Classifier.}
All CT series within a case are categorized by imaging region and contrast phase. Imaging region classification is performed using a LightGBM model trained on volume statistics derived from TotalSegmentator organ labels, categorizing series into head, neck, chest, abdomen, pelvis\cite{NIPS2017_6449f44a}. Performance details of the imaging region classifier are provided in Appendix~\ref{app:region_phase_classifier}. Contrast phase classification is performed using the TotalSegmentator contrast phase classification model (\texttt{totalseg\_get\_phase} entrypoint).

\textbf{Step 4: Finding––Series Matching.}
Each finding is then assigned to the most appropriate CT series within the case. Because the region and phase predicted by the Optimal Series Extractor may not exactly match the available CT series, a rule–based matching procedure is applied to select the best–fitting series. The detailed matching logic is described in Appendix~\ref{app:finding_series_matching}.

\textbf{Step 5: Organ Extractor.}
Finally, the Organ Extractor identifies the specific organ labels corresponding to each finding sentence by processing the finding text together with the available TotalSegmentator organ labels using the LLM (gpt-oss:20b). Findings for which no corresponding organ label can be confidently identified are excluded. The prompt used in this step is provided in Appendix~\ref{app:llm_prompts:organ_extractor}.

Through this fully automated pipeline, we generated approximately 224{,}000 organ–level volume––text pairs from 57{,}000 training cases. Detailed dataset statistics for each split, including the number of generated volume––text pairs, are summarized in Table~\ref{tab:dataset_statistics}.

\begin{table}[ht]
  \centering
  \caption{Dataset statistics for each split.}
  \label{tab:dataset_statistics}
  \begin{tabular}{p{4cm}cccc}
    \toprule
    Dataset                & VideoMAE Pretrain & Train      & Valid       & Test    \\
    \midrule
    Dates (2024/)          & 1/1––5/15          & 10/1––12/22 & 12/23––12/31 & 10/1––12/31 \\
    Institutes             & 8                 & 7          & 7           & 1         \\
    Patients               & 119,275           & 57,457     & 5,964       & 5,611     \\
    Series                 & 286,667           & 141,554    & 14,347      & 11,077    \\
    Reports                & ––                & 57,457     & 5,964       & 5,611     \\
    Volume–text pairs\newline (Original report sentence) & –– & 224,117    & 12,415      & 40,710    \\
    Volume–text pairs\newline (Rule–based negative)  & –– & 107,035    & 5,503       & 11,721    \\
    \bottomrule
  \end{tabular}
\end{table}

\subsection{Negative Data Augmentation}
Among the organ volume–text pairs of the training data generated by the aforementioned method, approximately 60\% were positive findings indicating the presence of a disease. Since radiologists primarily write reports to point out abnormalities, sentences indicating normalcy are rarely reflected in reports, and normal organs constitute the vast majority. Therefore, for the purpose of encouraging the learning of normal anatomy in contrastive learning and for data augmentation, we performed data augmentation of negative examples. Specifically, for organs that exist within a series but are not mentioned in the report, we judged them to be normal and created pairs as negative text by randomly selecting from multiple rule–based template sentences. Examples of templates include "No significant abnormality is observed in the \{organ\_name\}." As a result, as shown in Table \ref{tab:dataset_statistics}, approximately 110,000 negative pairs were added to the training data.

\subsection{Model Architecture and Training Details}
The base of our model architecture and training scheme is informed by InternVideo2 \cite{Wang2024-kk}. The image encoder is based on a 3D ViT, and optimizations were performed to efficiently and sufficiently encode CT volumes for each organ. The training scheme consists of three stages: pre–training with VideoMAE, volume–text contrastive learning, and VLM fine–tuning. Details of each step are shown below. Additionally, the data used and hyperparameters for each training stage are shown in Appendix \ref{app:training_details}.

\textbf{3D ViT Image Encoder.}
The image encoder has a 3D ViT structure with approximately the same parameter scale as ViT-B ($\approx$131M). The input size of the volume was set to 192 $\times$ 192 $\times$ 32, which is the average volume of the organs to be learned. For the H and W directions (axial plane), volumes smaller than 192 $\times$ 192 were expanded outward to match the target size, whereas larger volumes were resized to fit the target resolution. Along the Z direction (longitudinal axis), 32 consecutive slices were uniformly sampled. When the organ extent exceeded 32 slices, random sampling was applied during training, and a sliding–window strategy was used at inference to ensure full coverage. The patch size was 16 for H and W directions and 4 for the Z direction. For the purpose of accommodating various windowing conditions, windowing was performed with window levels and window widths for lung field conditions (-600, 1500), soft tissue conditions (40, 400), and bone conditions (300, 1500), respectively, normalized to 0––1 and input to the image encoder in 3 channels. All image tokens, including the cls token, were aggregated by attention pooling to produce a 768–dimensional image embedding.

\textbf{VideoMAE Pretraining.}
Before contrastive learning, the image encoder was pre–trained using VideoMAE, a self–supervised learning method, to acquire fundamental CT volume representations\cite{Tong2022-hq}. VideoMAE learns representations by dividing the input volume into 3D patches, randomly masking most of them, and reconstructing the masked patches using only the visible ones. In this study, the CT volume was treated as a 3D signal by interpreting the slice dimension as the temporal axis in VideoMAE. The masking rate was set to a uniform 75\%, and mean squared error (MSE) was used as the loss function. This VideoMAE pre–training improves learning stability in subsequent volume–text contrastive learning and facilitates the construction of visual representations robust to noise and variations in imaging conditions.

\textbf{Organ–separated Contrastive Learning.}
Following pre–training with VideoMAE, contrastive learning was conducted using Volume–Text pairs consisting of organ–segmented CT volumes and their corresponding radiology report sentences. The objective of this stage was to align organ–specific visual information with linguistic descriptions within a shared embedding space, explicitly enforcing organ–level semantic correspondence. We utilized a Japanese ModernBERT model (\texttt{sbintuitions/modernbert-ja-130m}) as the text encoder\cite{modernbert-ja}. The outputs from both the image and text encoders were projected into 768–dimensional embeddings through respective linear layers. After L2 normalization, each embedding was optimized using an InfoNCE loss based on cosine similarity, following the design principles of SigLIP \cite{Zhai2023-xw}.

\begin{figure}[ht]
  \centering
  \includegraphics[width=0.8\textwidth]{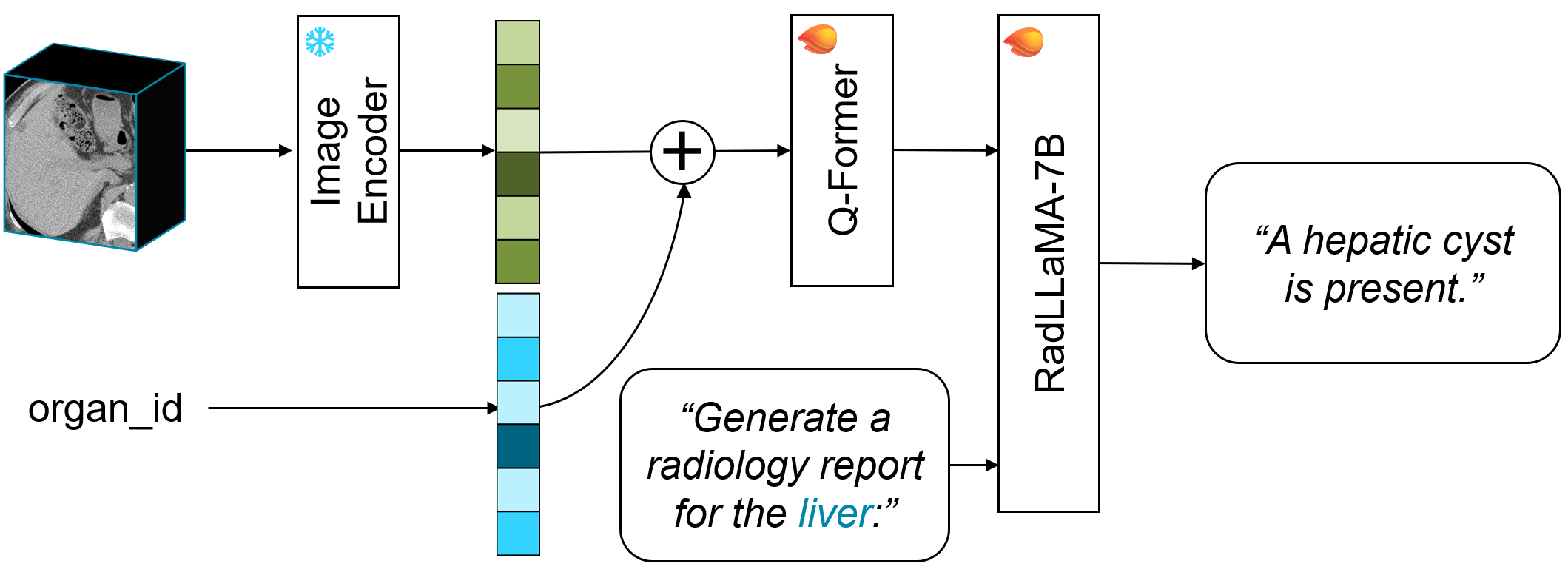}
  \caption{
    Overview of VLM Fine–tuning with Organ–specific Embeddings.
  }
  \label{fig:vlm_training_process}
\end{figure}

\textbf{VLM Fine–tuning.}
To compare VLM performance with Merlin, we utilized RadLLaMA-7B as the LLM component, consistent with the Merlin architecture \cite{Blankemeier2024-hk}. The Merlin dataset was employed for both training and evaluation purposes, following the predefined data splits provided by the dataset. A common approach for connecting an image encoder to an LLM is the Q-Former proposed in BLIP-2 \cite{Li2023-vr}. In this study, while we adopted the Q-Former as the adapter between the image encoder and the LLM, we implemented a custom training methodology. The original training scheme proposed for BLIP-2 consists of two stages: representation learning of visual information followed by alignment with the language space. This staggered approach is intended to improve stability by bridging the modality gap between image–text correspondence and language generation. In our framework, we mitigated this modality gap by simultaneously training the Q-Former and performing LoRA fine–tuning of the LLM, directly learning the connection to the second–stage LLM. Figure \ref{fig:vlm_training_process} provides an overview of our Q-Former training methodology for VLM construction. Organ–specific embeddings (\texttt{organ\_id}) were added to the embeddings output by the image encoder before being input to the Q-Former and subsequently connected to the LLM. During this process, the image encoder was frozen, and only the parameters for the Q-Former and the LLM's LoRA were trainable. The organ embeddings themselves were treated as trainable parameters. The Q-Former was initialized with weights from ``\texttt{Salesforce/blip2-opt-2.7b}''.

In Merlin, reports are generated at the regional level by specifying a region in the prompt and inputting the entire CT volume. In contrast, our VLM achieves more efficient and direct report generation by inputting only the specific organs identifiable from the target region. Because the regional labels in the Merlin dataset and the organ labels in TotalSegmentator do not always match perfectly, the mapping table used in this study is provided in Appendix \ref{app:merlin_totalseg_label_mapping}. In cases where a Merlin regional label (e.g., ``kidney'') corresponds to multiple organ labels (e.g., ``kidney\_left'' and ``kidney\_right''), all relevant organ embeddings were concatenated and input to the Q-Former, leveraging its ability to process variable–length image tokens. The prompts for the LLM followed the Merlin format, initiating report generation with the string: ``Generate a radiology report for \textless organ\_system\textgreater.''

\subsection{Evaluation}
To evaluate the zero–shot performance of TotalFM, we conducted two classification tasks: organ–wise lesion classification and finding–wise lesion classification. The former evaluates organ–level clinical findings using real radiology report sentences, while the latter follows a standard prompt–based zero–shot classification protocol for specific diseases. In addition, we assessed TotalFM as a VLM through quantitative radiology report generation metrics and qualitative human evaluation.

\textbf{Organ–wise lesion classification}
The volume–text pairs generated by our construction pipeline include labels indicating whether the corresponding finding sentence represents a positive or negative finding. Using these labels, we defined an organ–wise lesion classification task. Specifically, as illustrated in Fig. \ref{fig:research_overview}d, we established a binary classification task for each organ. Evaluation was performed by calculating the cosine similarity between the image and two text prompts (one indicating a positive finding and the other a negative finding) and classifying the image into the category with the higher similarity. Non–paired text prompts were generated according to the following rules:
1) If the pair data is positive: A rule–based template was used as the negative text (e.g., "No significant abnormality is observed in the \{organ\_name\}").
2) If the pair data is negative: A positive text for the same organ, randomly selected from the evaluation data, was used.
We conducted comparative evaluations against CT-CLIP \cite{Hamamci2024-xm} and Merlin \cite{Blankemeier2024-hk} for each organ. The F1 score was adopted as the primary evaluation metric, and positive and negative cases were strictly sampled at a 1:1 ratio for each organ.
For evaluation, TotalFM was provided with organ–cropped CT volumes and Japanese text inputs, whereas CT-CLIP and Merlin received full CT volumes and English prompts translated from the original Japanese sentences using an LLM (gpt-oss:20b).

\textbf{Finding–wise lesion classification}
For finding–wise lesion classification, we evaluated zero–shot performance on the J-MID and Merlin test datasets using AUROC. In J-MID, AUROC was computed for 30 major disease labels, while in Merlin, 17 localizable finding categories were evaluated. Similarity between the image embedding and a rule–based positive prompt (“{lesion} is present.”) was used as the confidence score. Merlin was used as the comparison model. In cases where a single disease label in TotalFM corresponded to multiple organ labels (e.g., kidney), the similarity between each organ and the text was calculated, and the average value was used as the final confidence score.

\textbf{Radiology Report Generation}
The evaluation of report generation followed the protocol and framework established by Merlin \cite{Blankemeier2024-hk}. Using Merlin as the baseline model, performance was assessed using four metrics: BLEU, ROUGE-2, BERTScore, and RadGraph-F1. Additionally, as a qualitative assessment, a professional evaluation of the generated reports was conducted by a radiologist with over 10 years of clinical experience. Detailed settings and results of this human evaluation are provided in Appendix \ref{app:human_eval}.
\section{Results}

\subsection{Organ–wise lesion classification}
Table \ref{tab:organ_wise_lesion_classification} presents the zero–shot lesion classification performance for each organ. The entries marked with "––" indicate organs that were not included in the respective model's training data; thus, evaluation metrics were not computed. Regarding the average F1 score for organs supported by both models, TotalFM achieved 0.708 compared to 0.515 for CT-CLIP, and 0.692 compared to 0.650 for Merlin. This represents an average performance improvement of 37.5\% over CT-CLIP and 6.5\% over Merlin. Our model achieved higher F1 scores in all organs except the aorta compared to CT-CLIP, and demonstrated superior performance in 64\% (9/14) of the organs compared to Merlin. Notably, TotalFM exhibited high generalization performance in organ–wise disease classification, with F1 scores exceeding 0.6 for all evaluated organ labels.

\begin{table}[htbp]
  \centering
  \caption{Comparison of zero–shot performance in organ–wise paired image–text lesion classification. The positive text corresponds to the original finding sentence from the CT report, while the negative text is a rule–based template indicating the absence of abnormality. F1 scores were computed based on whether the similarity with the positive text exceeded that with the negative one.}
  \label{tab:organ_wise_lesion_classification}
  \begin{tabular}{lccc}
    \toprule
              Organ &  CT-CLIP &  Merlin &  TotalFM (ours) \\
    \midrule
              brain &   ––   &   ––   &    .618 \\
      thyroid gland &   .354   &   ––   &    .731 \\
            trachea &   .478   &   ––   &    .709 \\
          esophagus &   .556   &   .662   &    .761 \\
               lung &    .475   &   .379   &    .612 \\
              aorta &   .650   &   .685   &    .611 \\
              heart &   .574   &   .694   &    .824 \\
              liver &   ––   &   .788   &    .739 \\
        gallbladder &   ––   &   .705   &    .694 \\
            stomach &   ––   &   .590   &    .666 \\
             spleen &   ––   &    .624   &    .704 \\
             kidney &   ––   &    .447   &    .708 \\
           pancreas &   ––   &    .666   &    .678 \\
        small bowel &   ––   &   .692   &    .749 \\
              colon &   ––   &   .767   &    .652 \\
    urinary bladder &   ––   &   .590   &    .618 \\
           prostate &   ––   &    .817   &    .675 \\
    \bottomrule
  \end{tabular}
\end{table}

\subsection{Finding–wise lesion classification}
\begin{figure}[ht]
  \centering
  \includegraphics[width=1.0\textwidth]{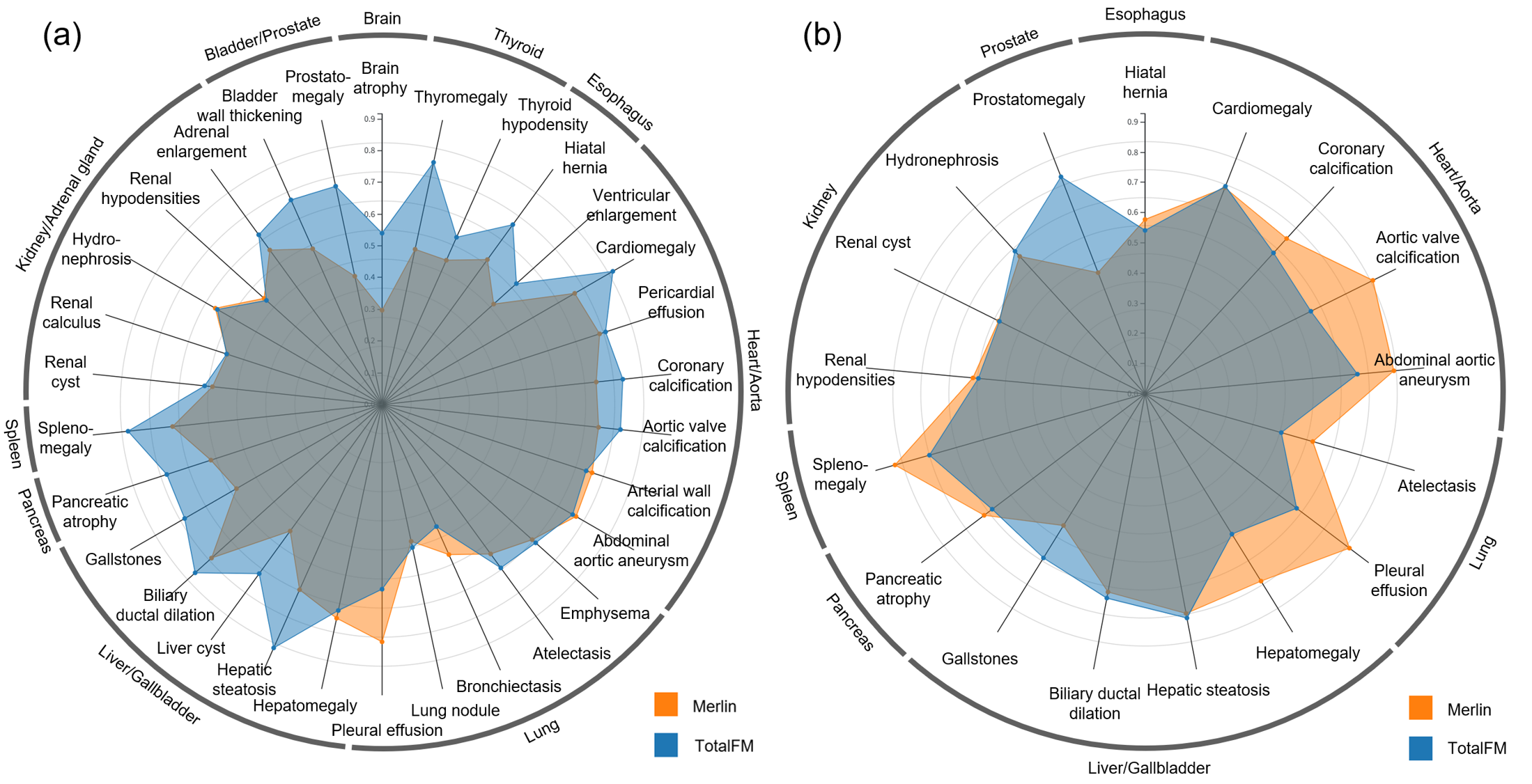}
  \caption{
    Comparison of AUROC performance by finding category. (a) AUROC comparison for 30 finding categories defined in the J-MID dataset. (b) AUROC comparison for 17 finding categories selected from the Merlin Test dataset.
  }
  \label{fig:finding_raderchart}
\end{figure}

Figure \ref{fig:finding_raderchart} illustrates the comparison of AUROC performance for each finding category using the J-MID dataset. Our model achieved a higher AUROC than Merlin in 83\% (25/30) of the finding categories. Conversely, for findings related to organs that are elongated along the z–axis (e.g., aorta, lungs) or organs that are divided into multiple TotalSegmentator labels, the performance was either comparable to or slightly lower than that of Merlin.

Furthermore, we conducted a benchmark evaluation using the Merlin Test dataset. We evaluated 17 out of the 30 finding categories provided in the Merlin dataset that could be localized using TotalSegmentator labels. In this benchmark as well, Merlin tended to show higher performance for findings related to the aorta and lung fields, while for other findings, the performance was comparable.

\subsection{Radiology Report Generation}
Table \ref{tab:radiology_report_generation} presents the quantitative evaluation of the radiology report generation results for each organ. Our model consistently outperformed Merlin across all evaluation metrics for the gallbladder and pelvic organs. For other organs, our model achieved report generation performance that was generally comparable to that of Merlin.

\begin{table}[htbp]
  \centering
  \caption{Quantitative evaluation of report generation results by organ.}
  \label{tab:radiology_report_generation}
  \begin{tabular}{lcccccccc}
    \toprule
    & \multicolumn{2}{c}{BLEU} & \multicolumn{2}{c}{ROUGE-2}  
    & \multicolumn{2}{c}{BERTScore} & \multicolumn{2}{c}{RadGraph-F1} \\
    \cmidrule(lr){2-3}\cmidrule(lr){4-5}\cmidrule(lr){6-7}\cmidrule(lr){8-9}
    Anatomical Region & Merlin & TotalFM & Merlin & TotalFM & Merlin & TotalFM & Merlin & TotalFM \\
    \midrule
    Lower thorax           & .043 & .032 & .270 & .265 & .394 & .314 & .346 & .306 \\
    Liver and biliary tree & .045 & .037 & .383 & .358 & .460 & .441 & .450 & .431 \\
    Gallbladder            & .228 & .233 & .552 & .604 & .750 & .776 & .626 & .725 \\
    Spleen                 & .338 & .338 & .709 & .709 & .826 & .827 & .823 & .823 \\
    Pancreas               & .116 & .110 & .704 & .706 & .802 & .804 & .766 & .767 \\
    Adrenal glands         & .508 & .508 & .862 & .862 & .907 & .907 & .898 & .898 \\
    Kidneys and ureters    & .017 & .051 & .387 & .348 & .476 & .444 & .417 & .385 \\
    Gastrointestinal tract & .034 & .024 & .125 & .120 & .234 & .266 & .248 & .208 \\
    Pelvic organs          & .028 & .139 & .114 & .167 & .183 & .219 & .216 & .265 \\
    \midrule
    Average                & .151 & .164 & .456 & .460 & .559 & .555 & .532 & .534 \\
    \bottomrule
  \end{tabular}
\end{table}

\section{Discussion}
In recent years, foundation models have achieved remarkable breakthroughs across diverse fields and have increasingly been explored as a key technology for comprehensive clinical support in radiology, including automated diagnosis and report generation\cite{Bannur2023-lf, Li2023-jl, Sellergren2025-hz}. However, developing foundation models for 3D volumetric data such as CT remains significantly more challenging compared to 2D imaging. This difficulty stems primarily from the spatial redundancy of the input data and high computational costs, which make it extremely difficult to secure sufficient batch sizes for large–scale pre–training and contrastive learning. Consequently, 3D–CT foundation models necessitate comprehensive optimization encompassing data structuring, model design, and training frameworks, rather than simple scaling.

In this study, we developed TotalFM, a foundation model trained on a large–scale CT database consisting of 280,000 series for pre–training and 140,000 series for contrastive learning. By constructing a Volume–Text Pair Construction Pipeline that leverages TotalSegmentator organ labels and prompt–based LLM processing, we generated organ–specific volume–text pairs. This enabled the model to learn the correspondence between CT images and linguistic expressions through an organ–separated contrastive learning framework. We believe this framework offers a promising and pragmatic approach for CT foundation models that utilize computationally expensive 3D volumetric data, allowing for the acquisition of superior representation capabilities even with limited computational resources in a contrastive learning setting where batch size directly impacts accuracy. Furthermore, in the VLM fine–tuning stage, we successfully mitigated the modality gap by combining Q–Former training with LoRA fine–tuning of the LLM, enabling direct learning of the connection to the LLM. Through these methodologies, our model achieved high zero–shot performance in both organ–wise and finding–wise lesion classification tasks, and demonstrated report generation performance comparable to Merlin. 

In the finding–wise classification task, TotalFM showed lower accuracy than Merlin for certain organs such as the aorta and lung fields. The aorta is a longitudinal organ elongated along the z–axis; in cases where the organ's z–dimension exceeded the model's input size, we employed a sliding window approach for inference. In this setting, the final prediction was calculated by averaging the similarities between multiple image embeddings and the text embedding. This averaging process likely diluted the signal for localized lesions, potentially leading to a decrease in sensitivity and overall accuracy. Similarly, the lung fields are divided into five labels (upper, middle, and lower lobes for both sides) in TotalSegmentator, and a similar dilution of local diagnostic information may have affected the results.

While finding–wise lesion classification has been frequently employed in previous research on 3D–CT foundation models, the organ–wise classification task using actual radiology report sentences—as proposed in this study—has remained largely unexplored. We consider this task to be a form of binary Visual Question Answering (VQA). By utilizing real–world radiology reports, we believe this approach allows for a more clinical evaluation of a foundation model's generalization capabilities. Furthermore, this method helps suppress performance fluctuations caused by subtle variations in prompt engineering, thereby contributing to a more intrinsic evaluation of a foundation model's performance. As several frameworks for evaluating 3D–CT foundation models have recently been proposed, our organ–wise classification task using actual report sentences stands as a robust candidate to contribute to the future standardization of foundation model evaluation\cite{Hamamci2024-xm, Agrawal2025-fy, hamamci2024developing}. By acquiring robust visual representations and image–text correspondences through VideoMAE and contrastive learning, TotalFM is expected to serve as a promising base model for diverse clinical applications. Performing inference for each organ present in a CT scan enables the construction of automated diagnostic systems for the entire volume. For instance, this could be implemented as an image–text retrieval system where text embeddings of finding sentences serve as a searchable index. Alternatively, when developing Computer–Aided Diagnosis (CAD) tools for detailed organ evaluation, TotalFM can be utilized as a foundation for fine–tuning, facilitating the creation of highly accurate diagnostic support systems.

Several limitations exist in this study. First, findings that do not correspond to the anatomical labels provided by TotalSegmentator were excluded. Specifically, findings related to the vascular system (excluding the aorta) and the lymphatic system were largely omitted from the training data. To address this, future dataset construction requires methods to localize regions within the CT volume that correspond to finding sentences, potentially utilizing techniques such as Open–Vocabulary Segmentation or Visual Grounding to learn richer local representations\cite{Zhao2025-tl, Ichinose2023-qe}. Second, while most previous studies perform contrastive learning between whole reports and whole volumes, our approach adopts a finer granularity by pairing specific finding sentences with organ volumes. A drawback of this approach is that the InfoNCE loss treats all non–paired elements in a mini–batch as negatives. This may introduce noise, as certain non–paired combinations within a batch might actually represent positive findings, potentially leading to over–separation of representations or unstable optimization \cite{Khosla2020-md, Kim2025-js}.Third, although we implemented a Series–Selector to choose a single representative CT series, clinical diagnosis often necessitates comparing multiple series to observe temporal changes, such as contrast enhancement dynamics. Future work should explore methods for extracting and simultaneously encoding multiple relevant series. Finally, while this study focused on individual findings, modeling the process of integrating these discrete findings into a comprehensive, holistic diagnosis remains a subject for future development.

\section{Conclusion}

We presented TotalFM, an organ–separated learning framework for 3D CT foundation models that combines self–supervised pre–training via VideoMAE with contrastive learning using organ–specific volume––text pairs. TotalFM achieves strong zero–shot performance in organ–wise and finding–wise lesion classification and demonstrates report generation performance comparable to existing vision––language models, while maintaining computational efficiency. These results indicate that organ–separated learning offers a practical solution to the trade–off between efficiency and representational capacity in 3D CT foundation models and provides a clinically meaningful basis for future downstream applications.

\textbf{Computational Hardware and Software}

Experiments were conducted in a Docker environment based on \texttt{pytorch/pytorch:2.6.0-cuda12.6-cudnn9-devel}. Key software packages and their versions are as follows: PyTorch 2.6.0, TorchVision 0.21.0, TorchAudio 2.6.0, Transformers 4.57.1, Accelerate 1.11.0, Datasets 4.3.0, huggingface-hub 0.36.0, Tokenizers 0.22.1, Safetensors 0.6.2, flash\_attn 2.8.3, open\_clip\_torch 3.2.0, timm 1.0.21, NumPy 2.2.2, SciPy 1.16.3, scikit-learn 1.7.2, pandas 2.3.3, and opencv-python 4.12.0.88. 
For LLM inference, we used the \texttt{gpt-oss:20b} model deployed via the Docker image \texttt{ollama/ollama:0.13.5}, with a temperature of 0.8 and a context length of 8192 tokens. 
Experiments were conducted on a server equipped with two NVIDIA H100 GPUs (96 GB VRAM each), 2 TB of system RAM, and 72 CPU cores.

\textbf{Data Availability}

The data used in this study are not publicly available due to restrictions related to patient privacy and ethical considerations. The datasets were accessed under approval from the relevant institutional ethics committee and are available from the corresponding author upon reasonable request, subject to institutional and ethical approval.

\textbf{Code Availability}

We plan to release the implementation code and pretrained model weights publicly in the future to facilitate transparency and reproducibility.

\textbf{Ethics Declaration}

This study was conducted in accordance with the Declaration of Helsinki and approved by the Jichi Medical University Hospital Bioethics Committee for Clinical Research (protocol code 25-087).


\begin{ack}
This research was partially supported by JSPS KAKENHI [Grant Number JP23K17234]. We would like to thank the departments of radiology that provided the J-MID database, including Juntendo Univ., Kyushu Univ., Keio Univ., The Univ. of Tokyo, Okayama Univ., Kyoto Univ., Osaka Univ., Tokyo Medical and Dental Univ., Hokkaido Univ., Ehime Univ., and Tokushima Univ.

\end{ack}


\medskip

{ 
\small
\bibliographystyle{ieeetr}
\bibliography{references}

@article{Xu2024-as,
  title     = {A whole-slide foundation model for digital pathology from
               real-world data},
  author    = {Xu, Hanwen and Usuyama, Naoto and Bagga, Jaspreet and Zhang,
               Sheng and Rao, Rajesh and Naumann, Tristan and Wong, Cliff and
               Gero, Zelalem and González, Javier and Gu, Yu and Xu, Yanbo and
               Wei, Mu and Wang, Wenhui and Ma, Shuming and Wei, Furu and Yang,
               Jianwei and Li, Chunyuan and Gao, Jianfeng and Rosemon, Jaylen
               and Bower, Tucker and Lee, Soohee and Weerasinghe, Roshanthi and
               Wright, Bill J and Robicsek, Ari and Piening, Brian and Bifulco,
               Carlo and Wang, Sheng and Poon, Hoifung},
  journal   = {Nature},
  publisher = {Springer Science and Business Media LLC},
  volume    = 630,
  number    = 8015,
  pages     = {181--188},
  month     = jun,
  year      = 2024,
  language  = {en}
}

@article{Wu2023-jg,
  title         = {Towards generalist foundation model for radiology by
                   leveraging web-scale {2D\&3D} medical data},
  author        = {Wu, Chaoyi and Zhang, Xiaoman and Zhang, Ya and Wang, Yanfeng
                   and Xie, Weidi},
  journal       = {arXiv [cs.CV]},
  month         = aug,
  year          = 2023,
  archiveprefix = {arXiv},
  primaryclass  = {cs.CV}
}

@article{Hamamci2024-xm,
  title         = {Developing generalist foundation models from a multimodal
                   dataset for {3D} computed tomography},
  author        = {Hamamci, Ibrahim Ethem and Er, Sezgin and Wang, Chenyu and
                   Almas, Furkan and Simsek, Ayse Gulnihan and Esirgun, Sevval
                   Nil and Doga, Irem and Durugol, Omer Faruk and Dai, Weicheng
                   and Xu, Murong and Dasdelen, Muhammed Furkan and Wittmann,
                   Bastian and Amiranashvili, Tamaz and Simsar, Enis and Simsar,
                   Mehmet and Erdemir, Emine Bensu and Alanbay, Abdullah and
                   Sekuboyina, Anjany and Lafci, Berkan and Bluethgen, Christian
                   and Batmanghelich, Kayhan and Ozdemir, Mehmet Kemal and
                   Menze, Bjoern},
  journal       = {arXiv [cs.CV]},
  month         = mar,
  year          = 2024,
  archiveprefix = {arXiv},
  primaryclass  = {cs.CV}
}

@article{Wang2024-kk,
  title         = {{InternVideo2}: Scaling foundation models for multimodal
                   video understanding},
  author        = {Wang, Yi and Li, Kunchang and Li, Xinhao and Yu, Jiashuo and
                   He, Yinan and Wang, Chenting and Chen, Guo and Pei, Baoqi and
                   Yan, Ziang and Zheng, Rongkun and Xu, Jilan and Wang, Zun and
                   Shi, Yansong and Jiang, Tianxiang and Li, Songze and Zhang,
                   Hongjie and Huang, Yifei and Qiao, Yu and Wang, Yali and
                   Wang, Limin},
  journal       = {arXiv [cs.CV]},
  month         = mar,
  year          = 2024,
  archiveprefix = {arXiv},
  primaryclass  = {cs.CV}
}

@article{Blankemeier2024-hk,
  title         = {Merlin: A vision language foundation model for {3D} computed
                   tomography},
  author        = {Blankemeier, Louis and Cohen, Joseph Paul and Kumar, Ashwin
                   and Van Veen, Dave and Gardezi, Syed Jamal Safdar and
                   Paschali, Magdalini and Chen, Zhihong and Delbrouck,
                   Jean-Benoit and Reis, Eduardo and Truyts, Cesar and
                   Bluethgen, Christian and Jensen, Malte Engmann Kjeldskov and
                   Ostmeier, Sophie and Varma, Maya and Valanarasu, Jeya Maria
                   Jose and Fang, Zhongnan and Huo, Zepeng and Nabulsi, Zaid and
                   Ardila, Diego and Weng, Wei-Hung and Junior, Edson Amaro and
                   Ahuja, Neera and Fries, Jason and Shah, Nigam H and Johnston,
                   Andrew and Boutin, Robert D and Wentland, Andrew and
                   Langlotz, Curtis P and Hom, Jason and Gatidis, Sergios and
                   Chaudhari, Akshay S},
  journal       = {arXiv [cs.CV]},
  month         = jun,
  year          = 2024,
  archiveprefix = {arXiv},
  primaryclass  = {cs.CV}
}

@article{Li2023-vr,
  title     = {{BLIP}-2: Bootstrapping language-image pre-training with frozen
               image encoders and large language models},
  author    = {Li, Junnan and Li, Dongxu and Savarese, S and Hoi, Steven C H},
  editor    = {Krause, Andreas and Brunskill, Emma and Cho, Kyunghyun and
               Engelhardt, Barbara and Sabato, Sivan and Scarlett, Jonathan},
  journal   = {ICML},
  publisher = {PMLR},
  volume    = 202,
  pages     = {19730--19742},
  month     = jan,
  year      = 2023
}

@misc{UnknownUnknown-at,
  title        = {Context-Aware Sentence Classification of Radiology Reports
                  Using Synthetic Data: Development and Validation Study},
  booktitle    = {JMIR Preprints},
  howpublished = {\url{https://preprints.jmir.org/preprint/86365}},
  note         = {Accessed: 2025-12-20},
  language     = {en}
}

@inproceedings{He2022-ti,
  title     = {Masked Autoencoders Are Scalable Vision Learners},
  author    = {He, Kaiming and Chen, Xinlei and Xie, Saining and Li, Yanghao and
               Dollár, Piotr and Girshick, Ross},
  booktitle = {Proceedings of the IEEE/CVF Conference on Computer Vision and
               Pattern Recognition},
  pages     = {16000--16009},
  year      = 2022
}

@inproceedings{Zhai2023-xw,
  title     = {Sigmoid Loss for Language Image Pre-Training},
  author    = {Zhai, Xiaohua and Mustafa, Basil and Kolesnikov, Alexander and
               Beyer, Lucas},
  booktitle = {Proceedings of the IEEE/CVF International Conference on Computer
               Vision},
  pages     = {11975--11986},
  year      = 2023
}

@article{Bai2024-nb,
  title         = {{M3D}: Advancing {3D} medical image analysis with multi-modal
                   large language models},
  author        = {Bai, Fan and Du, Yuxin and Huang, Tiejun and Meng, Max Q-H
                   and Zhao, Bo},
  journal       = {arXiv [cs.CV]},
  month         = mar,
  year          = 2024,
  archiveprefix = {arXiv},
  primaryclass  = {cs.CV}
}

@article{Zhou2022-qe,
  title         = {Self pre-training with masked autoencoders for medical image
                   classification and segmentation},
  author        = {Zhou, Lei and Liu, Huidong and Bae, Joseph and He, Junjun and
                   Samaras, Dimitris and Prasanna, Prateek},
  journal       = {arXiv [eess.IV]},
  month         = mar,
  year          = 2022,
  archiveprefix = {arXiv},
  primaryclass  = {eess.IV}
}

@article{van-den-Oord2018-pm,
  title         = {Representation learning with Contrastive Predictive Coding},
  author        = {van den Oord, Aaron and Li, Yazhe and Vinyals, Oriol},
  journal       = {arXiv [cs.LG]},
  month         = jul,
  year          = 2018,
  archiveprefix = {arXiv},
  primaryclass  = {cs.LG}
}

@inproceedings{Chen2020-du,
  title     = {A Simple Framework for Contrastive Learning of Visual
               Representations},
  author    = {Chen, Ting and Kornblith, Simon and Norouzi, Mohammad and Hinton,
               Geoffrey},
  booktitle = {International Conference on Machine Learning},
  publisher = {PMLR},
  pages     = {1597--1607},
  month     = nov,
  year      = 2020,
  language  = {en}
}

@article{Khosla2020-md,
  title         = {Supervised Contrastive Learning},
  author        = {Khosla, Prannay and Teterwak, Piotr and Wang, Chen and Sarna,
                   Aaron and Tian, Yonglong and Isola, Phillip and Maschinot,
                   Aaron and Liu, Ce and Krishnan, Dilip},
  journal       = {arXiv [cs.LG]},
  month         = apr,
  year          = 2020,
  archiveprefix = {arXiv},
  primaryclass  = {cs.LG}
}

@article{Kim2025-js,
  title         = {{FALCON}: False-negative Aware Learning of {COntrastive}
                   Negatives in vision-language alignment},
  author        = {Kim, Myunsoo and Shim, Seong-Woong and Lee, Byung-Jun},
  journal       = {arXiv [cs.CV]},
  month         = nov,
  year          = 2025,
  archiveprefix = {arXiv},
  primaryclass  = {cs.CV}
}

@article{Agrawal2025-fy,
  title         = {Pillar-0: A new frontier for radiology foundation models},
  author        = {Agrawal, Kumar Krishna and Liu, Longchao and Lian, Long and
                   Nercessian, Michael and Harguindeguy, Natalia and Wu, Yufu
                   and Mikhael, Peter and Lin, Gigin and Sequist, Lecia V and
                   Fintelmann, Florian and Darrell, Trevor and Bai, Yutong and
                   Chung, Maggie and Yala, Adam},
  journal       = {arXiv [cs.CV]},
  month         = nov,
  year          = 2025,
  archiveprefix = {arXiv},
  primaryclass  = {cs.CV},
  language      = {en}
}

@article{Wasserthal2023-mc,
  title     = {{TotalSegmentator}: Robust segmentation of 104 anatomic
               structures in {CT} images},
  author    = {Wasserthal, Jakob and Breit, Hanns-Christian and Meyer, Manfred T
               and Pradella, Maurice and Hinck, Daniel and Sauter, Alexander W
               and Heye, Tobias and Boll, Daniel T and Cyriac, Joshy and Yang,
               Shan and Bach, Michael and Segeroth, Martin},
  journal   = {Radiol. Artif. Intell.},
  publisher = {Radiological Society of North America},
  volume    = 5,
  number    = 5,
  pages     = {e230024},
  month     = sep,
  year      = 2023,
  language  = {en}
}

@inproceedings{Radford2021-ro,
  title     = {Learning Transferable Visual Models From Natural Language
               Supervision},
  author    = {Radford, Alec and Kim, Jong Wook and Hallacy, Chris and Ramesh,
               Aditya and Goh, Gabriel and Agarwal, Sandhini and Sastry, Girish
               and Askell, Amanda and Mishkin, Pamela and Clark, Jack and
               Krueger, Gretchen and Sutskever, Ilya},
  booktitle = {International Conference on Machine Learning},
  publisher = {PMLR},
  pages     = {8748--8763},
  month     = jul,
  year      = 2021,
  language  = {en}
}

@article{Tong2022-hq,
  title         = {{VideoMAE}: Masked autoencoders are data-efficient learners
                   for self-supervised video pre-training},
  author        = {Tong, Zhan and Song, Yibing and Wang, Jue and Wang, Limin},
  editor        = {Koyejo, S and Mohamed, S and Agarwal, A and Belgrave, D and
                   Cho, K and Oh, A},
  journal       = {arXiv [cs.CV]},
  pages         = {10078--10093},
  month         = mar,
  year          = 2022,
  archiveprefix = {arXiv},
  primaryclass  = {cs.CV}
}

@article{Akashi2025-xu,
  title     = {Japan-Medical Image Database ({J}-{MID}): Medical big data
               supporting data science},
  author    = {Akashi, Toshiaki and Kumamaru, Kanako K and Wada, Akihiko and
               Hashimoto, Masahiro and Hirata, Kenji and Hayakawa, Yayoi and
               Sano, Katsuhiro and Kamagata, Koji and Hagiwara, Akifumi and
               Ikenouchi, Yutaka and Aoki, Shigeki},
  journal   = {Juntendo Med. J.},
  publisher = {The Jutendo Medical Journal},
  volume    = 71,
  number    = 3,
  pages     = {166--172},
  month     = jun,
  year      = 2025,
  language  = {en}
}

@article{Touvron2023-si,
  title         = {{LLaMA}: Open and efficient foundation language models},
  author        = {Touvron, Hugo and Lavril, Thibaut and Izacard, Gautier and
                   Martinet, Xavier and Lachaux, Marie-Anne and Lacroix,
                   Timothée and Rozière, Baptiste and Goyal, Naman and Hambro,
                   Eric and Azhar, Faisal and Rodriguez, Aurelien and Joulin,
                   Armand and Grave, Edouard and Lample, Guillaume},
  journal       = {arXiv [cs.CL]},
  month         = feb,
  year          = 2023,
  archiveprefix = {arXiv},
  primaryclass  = {cs.CL}
}

@inproceedings{Bannur2023-lf,
  title     = {Learning To Exploit Temporal Structure for Biomedical
               Vision-Language Processing},
  author    = {Bannur, Shruthi and Hyland, Stephanie and Liu, Qianchu and
               Pérez-García, Fernando and Ilse, Maximilian and Castro, Daniel C
               and Boecking, Benedikt and Sharma, Harshita and Bouzid, Kenza and
               Thieme, Anja and Schwaighofer, Anton and Wetscherek, Maria and
               Lungren, Matthew P and Nori, Aditya and Alvarez-Valle, Javier and
               Oktay, Ozan},
  booktitle = {Proceedings of the IEEE/CVF Conference on Computer Vision and
               Pattern Recognition},
  pages     = {15016--15027},
  year      = 2023
}

@article{Li2023-jl,
  title         = {{LLaVA}-Med: Training a Large Language-and-Vision Assistant
                   for {BioMedicine} in one day},
  author        = {Li, Chunyuan and Wong, Cliff and Zhang, Sheng and Usuyama,
                   Naoto and Liu, Haotian and Yang, Jianwei and Naumann, Tristan
                   and Poon, Hoifung and Gao, Jianfeng},
  editor        = {Oh, A and Naumann, T and Globerson, A and Saenko, K and
                   Hardt, M and Levine, S},
  journal       = {arXiv [cs.CV]},
  pages         = {28541--28564},
  month         = jun,
  year          = 2023,
  archiveprefix = {arXiv},
  primaryclass  = {cs.CV}
}

@article{Sellergren2025-hz,
  title         = {{MedGemma} Technical Report},
  author        = {Sellergren, Andrew and Kazemzadeh, Sahar and Jaroensri, Tiam
                   and Kiraly, Atilla and Traverse, Madeleine and Kohlberger,
                   Timo and Xu, Shawn and Jamil, Fayaz and Hughes, Cían and Lau,
                   Charles and Chen, Justin and Mahvar, Fereshteh and Yatziv,
                   Liron and Chen, Tiffany and Sterling, Bram and Baby, Stefanie
                   Anna and Baby, Susanna Maria and Lai, Jeremy and Schmidgall,
                   Samuel and Yang, Lu and Chen, Kejia and Bjornsson, Per and
                   Reddy, Shashir and Brush, Ryan and Philbrick, Kenneth and
                   Asiedu, Mercy and Mezerreg, Ines and Hu, Howard and Yang,
                   Howard and Tiwari, Richa and Jansen, Sunny and Singh, Preeti
                   and Liu, Yun and Azizi, Shekoofeh and Kamath, Aishwarya and
                   Ferret, Johan and Pathak, Shreya and Vieillard, Nino and
                   Merhej, Ramona and Perrin, Sarah and Matejovicova, Tatiana
                   and Ramé, Alexandre and Riviere, Morgane and Rouillard, Louis
                   and Mesnard, Thomas and Cideron, Geoffrey and Grill,
                   Jean-Bastien and Ramos, Sabela and Yvinec, Edouard and
                   Casbon, Michelle and Buchatskaya, Elena and Alayrac,
                   Jean-Baptiste and Lepikhin, Dmitry and Feinberg, Vlad and
                   Borgeaud, Sebastian and Andreev, Alek and Hardin, Cassidy and
                   Dadashi, Robert and Hussenot, Léonard and Joulin, Armand and
                   Bachem, Olivier and Matias, Yossi and Chou, Katherine and
                   Hassidim, Avinatan and Goel, Kavi and Farabet, Clement and
                   Barral, Joelle and Warkentin, Tris and Shlens, Jonathon and
                   Fleet, David and Cotruta, Victor and Sanseviero, Omar and
                   Martins, Gus and Kirk, Phoebe and Rao, Anand and Shetty,
                   Shravya and Steiner, David F and Kirmizibayrak, Can and
                   Pilgrim, Rory and Golden, Daniel and Yang, Lin},
  journal       = {arXiv [cs.AI]},
  month         = jul,
  year          = 2025,
  archiveprefix = {arXiv},
  primaryclass  = {cs.AI}
}

@misc{modernbert-ja,
  author       = {Tsukagoshi, Hayato and Li, Shengzhe and Fukuchi, Akihiko and Shibata, Tomohide},
  title        = {{ModernBERT-Ja}},
  howpublished = {\url{https://huggingface.co/collections/sbintuitions/modernbert-ja-67b68fe891132877cf67aa0a}},
  url          = {https://huggingface.co/collections/sbintuitions/modernbert-ja-67b68fe891132877cf67aa0a},
  year         = {2025}
}

@misc{openai2025gptoss120bgptoss20bmodel,
  title         = {gpt-oss-120b \& gpt-oss-20b Model Card},
  author        = {OpenAI},
  year          = {2025},
  eprint        = {2508.10925},
  archiveprefix = {arXiv},
  primaryclass  = {cs.CL},
  url           = {https://arxiv.org/abs/2508.10925}
}

@inproceedings{NIPS2017_6449f44a,
  author    = {Ke, Guolin and Meng, Qi and Finley, Thomas and Wang, Taifeng and Chen, Wei and Ma, Weidong and Ye, Qiwei and Liu, Tie-Yan},
  booktitle = {Advances in Neural Information Processing Systems},
  editor    = {I. Guyon and U. Von Luxburg and S. Bengio and H. Wallach and R. Fergus and S. Vishwanathan and R. Garnett},
  pages     = {},
  publisher = {Curran Associates, Inc.},
  title     = {LightGBM: A Highly Efficient Gradient Boosting Decision Tree},
  url       = {https://proceedings.neurips.cc/paper_files/paper/2017/file/6449f44a102fde848669bdd9eb6b76fa-Paper.pdf},
  volume    = {30},
  year      = {2017}
}

@article{hamamci2024developing,
  title   = {Developing Generalist Foundation Models from a Multimodal Dataset for 3D Computed Tomography},
  author  = {Hamamci, Ibrahim Ethem and Er, Sezgin and Almas, Furkan and Simsek, Ayse Gulnihan and Esirgun, Sevval Nil and Dogan, Irem and Dasdelen, Muhammed Furkan and Durugol, Omer Faruk and Wittmann, Bastian and Amiranashvili, Tamaz and others},
  journal = {arXiv preprint arXiv:2403.17834},
  year    = {2024}
}

@article{Zhao2025-tl,
  title     = {Large-vocabulary segmentation for medical images with text
               prompts},
  author    = {Zhao, Ziheng and Zhang, Yao and Wu, Chaoyi and Zhang, Xiaoman and
               Zhou, Xiao and Zhang, Ya and Wang, Yanfeng and Xie, Weidi},
  journal   = {NPJ Digit. Med.},
  publisher = {Springer Science and Business Media LLC},
  volume    = 8,
  number    = 1,
  pages     = 566,
  month     = sep,
  year      = 2025,
  language  = {en}
}

@incollection{Ichinose2023-qe,
  title     = {Visual grounding of whole radiology reports for {3D} {CT} images},
  author    = {Ichinose, Akimichi and Hatsutani, Taro and Nakamura, Keigo and
               Kitamura, Yoshiro and Iizuka, Satoshi and Simo-Serra, Edgar and
               Kido, Shoji and Tomiyama, Noriyuki},
  booktitle = {Lecture Notes in Computer Science},
  publisher = {Springer Nature Switzerland},
  address   = {Cham},
  pages     = {611--621},
  series    = {Lecture Notes in Computer Science},
  year      = 2023,
  language  = {en}
}
}

\clearpage
\appendix

\section{Radiology Reports Generated by the VLM and Human Evaluation by a Radiologist}\label{app:human_eval}
To evaluate the clinical validity and descriptive quality of the generated radiology reports, we conducted a human evaluation. This assessment was performed by a single radiologist with over 10 years of clinical experience. The objective of this human evaluation was to qualitatively analyze aspects that automated evaluation metrics often fail to capture, such as the accuracy of the descriptive content from a clinical perspective, the presence of omissions or redundancies, and any potentially misleading expressions.

\begin{figure}[ht]
  \centering
  \includegraphics[width=1.0\textwidth]{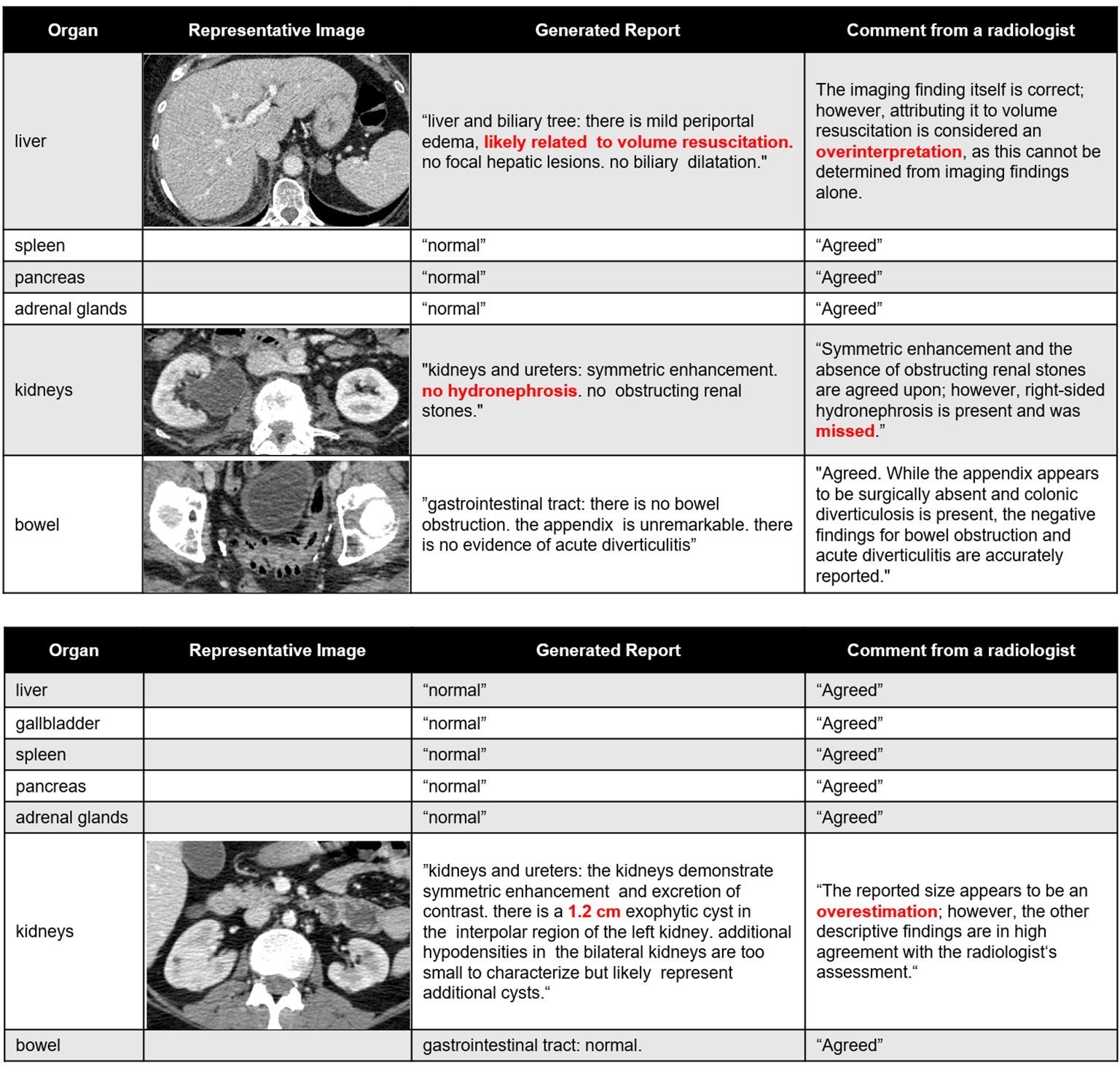}
  \caption{
    Examples of organ–specific radiology reports generated by TotalFM (VLM) and their corresponding human evaluations by a radiologist.
  }
  \label{fig:human_eval}
\end{figure}

Figure \ref{fig:human_eval} presents two representative examples of this human evaluation. For each case, we provide a representative image of the target organ, the corresponding model–generated report, and the associated comments from the radiologist. Within the generated reports and the radiologist's comments, descriptions judged to be clinically inaccurate or inappropriate are highlighted in red.

\section{Detailed Dataset Statistics}
Table \ref{tab:dataset_range_statistics} summarizes the imaging ranges of the dataset collected from J-MID used in this study.

\begin{table}[ht]
  \centering
  \caption{Distribution of CT imaging regions in the dataset.}
  \label{tab:dataset_range_statistics}
  \begin{tabular}{lccc}
    \toprule
    Imaging Regions    & Train    & Valid   & Test    \\
    \midrule
    Whole body      & 15,606   & 1,726   & 1,142   \\
    Trunk           & 25,230   & 2,608   & 2,692   \\
    Head            & 6,321    & 572     & 661     \\
    Chest           & 64,612   & 5,916   & 3,840   \\
    Abdomen/Pelvis  & 17,719   & 1,946   & 1,729   \\
    Other           & 12,066   & 1,353   & 1,013   \\
    \bottomrule
  \end{tabular}
\end{table}

\section{LLM Prompts}\label{app:llm_prompts}

\subsection{Optimal Series Extractor Prompt}\label{app:llm_prompts:optimal_series_extractor}
The prompt used for the Optimal Series Extractor, which classifies the optimal imaging range and contrast phase from a finding sentence, is shown below. Note that gpt-oss:20b was used as the actual LLM. While the actual prompt was written in Japanese (the same language as the original radiology reports), the version provided here is a translation for the purpose of this paper:

\begin{Verbatim}[commandchars=\\\{\}]
\textbf{[System Prompt]}
You are an experienced radiologist. Based on the following guidelines, determine which 
imaging range the given CT finding appears in and the appropriate contrast phase, 
then report it in JSON format.

{
  "range": "\textit{<Imaging Range>}",
  "phase": "\textit{<Contrast Phase>}"
}

Choices for "range":
  - Head
  - Neck
  - Chest (Lung)
  - Chest (Non-lung)
  - Abdomen
  - Pelvis
  - Other

Examples:
  - Lung nodule -> "range": "Chest (Lung)"
  - Cardiomegaly -> "range": "Chest (Non-lung)"
  - Abdominal aortic aneurysm -> "range": "Abdomen"

* Use "Other" if the finding does not fall into any of the above categories or is 
unclear (e.g., "No other abnormalities observed").

Choices for "phase":
  - Non-contrast
  - Early arterial phase
  - Late arterial phase
  - Portal phase or later
  - Not specified

* Generally, you may select "Not specified," but if the finding implies observation 
during a specific contrast phase, select the appropriate one.
* Please strictly adhere to the provided choices.

\textbf{[User Prompt]}
\textit{<Finding sentence>}
\end{Verbatim}

\subsection{Organ Extractor Prompt}\label{app:llm_prompts:organ_extractor}
The prompt used for the Organ Extractor, which identifies specific organ labels from a provided list based on the finding sentence within a Series–Text pair, is shown below. Similar to the other pipeline components, gpt-oss:20b was utilized as the LLM. The actual prompt was conducted in Japanese (the same language as the radiology reports), and the version provided here is a translation for reference:

\begin{Verbatim}[commandchars=\\\{\}]
\textbf{[System Prompt]}
You are an expert in medical image analysis. Please identify the relevant organ 
labels from the given medical finding sentence.

Instructions:
1. Analyze the finding sentence and identify the mentioned organs.
2. Select all applicable organ labels from the provided list and respond 
   in a list format.
3. Output only the organ label names (no explanations or additional text).
4. If no corresponding organ is found or the confidence is low, respond 
   with ["unknown"].

## Example 1
Finding sentence: "Cardiac enlargement is observed on chest X-ray."
Output: ["heart"]
## Example 2
Finding sentence: "Emphysematous changes in both lungs."
Output: ["lung_upper_lobe_left", "lung_lower_lobe_left", "lung_upper_lobe_right", 
"lung_middle_lobe_right", "lung_lower_lobe_right"]
## Example 3
Finding sentence: "No particular abnormalities are observed."
Output: ["unknown"]

\textbf{[User Prompt]}
Finding sentence: \textit{<finding_text>}

Available organ labels: \textit{<organs_list>}
\end{Verbatim}

\section{Performance of the Imaging Region Classifier}
\label{app:region_phase_classifier}

We report the performance of the imaging region classifier used for categorizing CT series into major anatomical regions. The classifier was evaluated using the area under the receiver operating characteristic curve (AUROC).

\begin{table}[ht]
  \centering
  \caption{AUROC of the imaging region classifier for major anatomical regions.}
  \label{tab:region_classifier_auc}
  \begin{tabular}{lc}
    \toprule
    Imaging region & AUROC \\
    \midrule
    Head    & 0.9974 \\
    Neck    & 0.9654 \\
    Chest   & 0.9708 \\
    Abdomen & 0.9897 \\
    Pelvis  & 0.9839 \\
    \bottomrule
  \end{tabular}
\end{table}

\section{Training Details}\label{app:training_details}
The hyperparameters used in each training process are summarized in Table \ref{tab:hyperparams}.

\begin{table}[t]
  \centering
  \caption{Hyperparameters used for each training stage.}
  \label{tab:hyperparams}
  \begin{tabular}{lccc}
    \toprule
    Parameters & VideoMAE Pretrain & Contrastive Learning & VLM Fine–tuning \\
    \midrule
    Optimizer & AdamW & AdamW & AdamW \\
    Base learning rate & $1.5\times10^{-4}$ & $1.0\times10^{-4}$ & $5.0\times10^{-5}$ \\
    Weight decay & 0.05 & $1.0\times10^{-4}$ & 0.01 \\
    Optimizer momentum & $\beta_1=0.9,\ \beta_2=0.999$ & $\beta_1=0.9,\ \beta_2=0.999$ & $\beta_1=0.9,\ \beta_2=0.999$ \\
    Batch size & 64 & 64 & 32 \\
    Training epochs & 20 & 24 & 20 \\
    Learning rate schedule & Cosine decay & Cosine decay & Cosine decay \\
    Max text length & –– & 512 & 512 \\
    Gradient accumulation & 2 & 2 & None \\
    Gradient clipping & 1.0 & 1.0 & 1.0 \\
    \bottomrule
  \end{tabular}
\end{table}

\section{Finding–Series Matching}\label{app:finding_series_matching}
The logic for matching CT series with finding sentences, based on the imaging range and contrast phase classified by the Optimal Series Extractor from the finding text, is as follows:

\begin{algorithm}[H]
\DontPrintSemicolon
\caption{Representative CT Series Selection}
\label{alg:rep_ct_series_selection}

\KwIn{$\mathcal{C}$: set of all available series in a study}
\KwIn{$R_{\text{target}}$: target imaging region (from the finding)}
\KwIn{$\Phi_{\text{target}}$: target contrast phase (from the finding)}
\KwOut{$s_{\text{rep}}$: a single representative series}

\BlankLine
\textbf{(1) Region filtering}\;
$C \leftarrow \{\, s \in \mathcal{C} \mid \texttt{is\_match}(s.\texttt{region}, R_{\text{target}})\,\}$\;

\BlankLine
\textbf{(2) Reconstruction style filtering}\;
\uIf{$R_{\text{target}}$ is \texttt{Lung}}{
  $C_{\text{tmp}} \leftarrow \{\, s \in C \mid (s.\texttt{kernel} \text{ is \texttt{Lung}})\ \lor\ (s.\texttt{WW} \ge 1000)\,\}$\;
}
\Else{
  $C_{\text{tmp}} \leftarrow \{\, s \in C \mid s.\texttt{kernel} \text{ is \texttt{SoftTissue}}\,\}$\;
}
\If{$C_{\text{tmp}} \neq \emptyset$}{
  $C \leftarrow C_{\text{tmp}}$\;
}

\BlankLine
\textbf{(3) Phase prioritization}\;
\uIf(\tcp*[f]{Phase specified}){$\Phi_{\text{target}}$ is specified}{
  $C \leftarrow \{\, s \in C \mid s.\texttt{phase} = \Phi_{\text{target}}\,\}$\;
}
\Else(\tcp*[f]{No phase specified}) {
  \tcp{Prefer Portal $\succ$ Late-Arterial $\succ$ Native $\succ$ Early-Arterial}
  $C \leftarrow \texttt{top\_ranked\_group}(C)$\;
}

\BlankLine
\textbf{(4) Deterministic tie–breaking}\;
\uIf{$|C| > 1$}{
  \Return $\,\arg\min_{s \in C}\ s.\texttt{SeriesNumber}$\;
}
\Else{
  \Return the remaining $s \in C$\;
}

\end{algorithm}

\section{Mapping Table between Merlin Anatomical Labels and TotalSegmentator Organ Labels}\label{app:merlin_totalseg_label_mapping}
During VLM training, as the anatomical labels defined in the Merlin dataset and the organ labels from TotalSegmentator do not always perfectly align, the mapping table used in this study is presented in Table \ref{tab:merlin_totalseg_label_mapping}. Furthermore, cases where specific organ labels could not be identified from the Merlin anatomical labels were excluded from both training and evaluation.
\begin{table}[htbp]
  \centering
  \caption{Mapping table between Merlin anatomical region labels and TotalSegmentator organ labels used in VLM training.}
  \label{tab:merlin_totalseg_label_mapping}
  \begin{tabular}{p{8cm}c}
    \toprule
    Merlin Anatomical Region & TotalSegmentator Organ Labels \\
    \midrule
    lower thorax | lower chest | lung bases & lung\_lower\_lobe\_left, lung\_lower\_lobe\_right \\
    liver | liver and biliary tree | biliary system & liver \\
    gallbladder & gallbladder \\
    spleen & spleen \\
    pancreas & pancreas \\
    adrenal glands | adrenals & adrenal\_gland\_right, adrenal\_gland\_left \\
    kidneys | kidneys and ureters | gu | kidneys, ureters & kidney\_right, kidney\_left \\
    bowel | gastrointestinal tract | gi | bowel/mesentery & small\_bowel, duodenum, colon \\
    pelvic organs | bladder | prostate and seminal vesicles | pelvis | uterus and ovaries & urinary\_bladder, prostate \\
    vasculature & heart, aorta \\
    \bottomrule
  \end{tabular}
\end{table}

\end{document}